\documentclass[sigconf=true, nonacm=True, review=false, anonymous = false,screen=true]{acmart}

\usepackage{dsfont}

\AtBeginDocument{%
  \providecommand\BibTeX{{%
    \normalfont B\kern-0.5em{\scshape i\kern-0.25em b}\kern-0.8em\TeX}}}

\copyrightyear{2023} 
\acmYear{2023} 
\setcopyright{rightsretained} 
\acmConference[GECCO '23]{Genetic and Evolutionary Computation Conference}{July 15--19, 2023}{Lisbon, Portugal}
\acmBooktitle{Genetic and Evolutionary Computation Conference (GECCO '23), July 15--19, 2023, Lisbon, Portugal}\acmDOI{10.1145/3583131.3590410}
\acmISBN{979-8-4007-0119-1/23/07}

\begin{document}

\title[When to be Discrete]{When to be Discrete: Analyzing Algorithm Performance on Discretized Continuous Problems}

\author{André Thomaser}
\affiliation{%
  \institution{BMW Group}
  \streetaddress{Knorrstraße 147}
  \city{Munich}
  \postcode{80788}
  \country{Germany}}
\email{andre.thomaser@bmw.de}
\orcid{0000-0002-6210-8784}

\author{Jacob de Nobel}
\affiliation{%
  \institution{LIACS, Leiden University}
  \streetaddress{Niels Bohrweg 1}
  \city{Leiden}
  \postcode{2333}
  \country{The Netherlands}}
\email{j.p.de.nobel@liacs.leidenuniv.nl}
\orcid{0000-0003-1169-1962}

\author{Diederick Vermetten}
\affiliation{%
  \institution{LIACS, Leiden University}
  \streetaddress{Niels Bohrweg 1}
  \city{Leiden}
  \postcode{2333}
  \country{The Netherlands}}
\email{d.l.vermetten@liacs.leidenuniv.nl}
\orcid{0000-0003-3040-7162}

\author{Furong Ye}
\affiliation{%
  \institution{LIACS, Leiden University}
  \streetaddress{Niels Bohrweg 1}
  \city{Leiden}
  \postcode{2333}
  \country{The Netherlands}}
\email{f.ye@liacs.leidenuniv.nl}
\orcid{0000-0002-8707-4189}

\author{Thomas Bäck}
\affiliation{%
  \institution{LIACS, Leiden University}
  \streetaddress{Niels Bohrweg 1}
  \city{Leiden}
  \postcode{2333}
  \country{The Netherlands}}
\email{t.h.w.baeck@liacs.leidenuniv.nl}
\orcid{0000-0001-6768-1478}

\author{Anna V. Kononova}
\affiliation{%
  \institution{LIACS, Leiden University}
  \streetaddress{Niels Bohrweg 1}
  \city{Leiden}
  \postcode{2333}
  \country{The Netherlands}}
\email{a.kononova@liacs.leidenuniv.nl}
\orcid{0000-0002-4138-7024}


\begin{abstract}

The domain of an optimization problem is seen as one of its most important characteristics. In particular, the distinction between continuous and discrete optimization is rather impactful. Based on this, the optimizing algorithm, analyzing method, and more are specified. However, in practice, no problem is ever truly continuous. Whether this is caused by computing limits or more tangible properties of the problem, most variables have a finite resolution. \\

In this work, we use the notion of the resolution of continuous variables to discretize problems from the continuous domain. We explore how the resolution impacts the performance of continuous optimization algorithms. Through a mapping to integer space, we are able to compare these continuous optimizers to discrete algorithms on the exact same problems. We show that the standard $(\mu_W, \lambda)$-CMA-ES fails when discretization is added to the problem. 


  
\end{abstract}

\ccsdesc[500]{Theory of computation~Discrete optimization}
\ccsdesc[300]{Theory of computation~Evolutionary algorithms}
\ccsdesc[300]{Theory of computation~Continuous optimization}

\keywords{Discretization, benchmarking, CMA-ES, integer representation, integer handling}

\maketitle

\section{Introduction}
The way in which optimization problems represent their domain has been a topic of debate in evolutionary computation since its beginning~\citep{Whitley2012}. In the early years of the field, this notion of representation divided the Genetic Algorithms (GA), which used binary representation, from Evolutionary Strategies (ES), which relied on the continuous encoding of variables. While many works in the 1990s managed to close this gap and bring the communities together~\citep{Baeck.1993}, the separation between discrete and continuous optimization is still present. 
This seems logical, as problem representation directly influences the available algorithms and analysis tools, and has many theoretical implications~\citep{Whitley2012,Rudolph2012}. Nonetheless, in practice, the distinction between a continuous and a discrete variable is oftentimes not as clear as it might at first appear. For example, in the case of optimization of an industrial design, considerations such as a physical limit on the actually attainable precision of the optimized variables come into play. On the other hand, continuous variables are typically represented in computation as floating point numbers, which have finite precision and are known to lead to numerical instabilities in some extreme cases~\citep{Mesnard2017}.   \\

In order to better understand the impact of the choice of representation on the performance of optimization algorithms, we first formalize the problem setting. In single-objective black-box optimization setting, the goal is to find a solution~$\mathbf{x}^*$ within a feasible set~$\mathcal{X}$ that minimizes an objective function~$f(x): \mathcal{X} \mapsto \mathbb{R}$:
\begin{equation}
\mathbf{x}^* = \arg \min_{x\in \mathcal{X}} f(x)
\end{equation}
In black-box optimization, the analytical form of the objective function~$f$ is unknown to the solver and treated as a black box that in response to an input provides a (continuous) output~\cite{Audet.2017}. Often, real-world optimization problems can be categorized as black-box optimization problems, as they frequently involve some (modeled) process or simulation~\citep{Long.2022, Thomaser.2022}, for which the exact mathematical definition is unknown or hard to access. Furthermore, such optimization problems are often categorized into continuous and discrete problems~\cite{Audet.2017}, where for \textit{continuous} problems $\mathcal{X} \subseteq \mathbb{R}^n$, while the feasible set for \textit{discrete} optimization problems is defined by $\mathcal{X} \subseteq \mathbb{Z}^n$. Combinations of both discrete and continuous variables are considered by the \textit{mixed-integer} domain~\citep{Floudas.1995}. \\

Although numerous studies have been conducted for continuous and discrete optimization domains, unfortunately, the researcher from the two domain communities usually focus on the separate domain without efficient communication, resulting in a gap of different techniques between the two domains. This issue has been widely discussed in the EC community~\cite{DBLP:journals/corr/abs-2007-03488}. However, this separation of research knowledge between two domains is a barrier for non-expert users. As a result, many variants of the same algorithm have been proposed for both continuous and discrete domains~\cite{DBLP:conf/gecco/PanTL07,DBLP:journals/swevo/MirjaliliL13,fan2015self,deep2009real}. While we are not critical of those algorithms, it is interesting to study how similar techniques can be transmitted from one type of search space to the other. Also, another question is how the problem variables impact those commonly used algorithm techniques. Regarding the transition of variable types, the problem suites from the famous 2005 CEC competition in real parameter optimization have been applied as a benchmark suite for discrete optimization methods~\cite{DBLP:journals/swevo/MirjaliliL13,fan2015self}. \\

In this study, we consider the case roughly inspired by an industrial application~\citep{Thomaser.2022}, where continuous input variables are defined up to a certain precision, which essentially transforms it into a discrete problem. 
As discussed above, many algorithms exist in discrete and continuous optimization domains. While searching for a proper algorithm to solve our problem, we aim to study the point at which it becomes beneficial to use either a continuous or a discrete algorithm when the limiting precision is varied. We introduce the notion of \textit{resolution} $\epsilon$, which we define as the precision of the floating point encoding of the continuous value. In this sense, with a higher resolution, the representation of the problem approaches a true continuous form, while for lower levels of resolution, the number of discrete levels in the representation reduces. Additionally, we define the concept of \textit{plateau size} $\rho$, which can be thought of as the reciprocal of the resolution, or the size of the discretization step. Conversely, a large plateau size corresponds to a low resolution while smaller plateau sizes increase the number of discrete levels in the representation. \\  

In this work, we focus on the relationship between the resolution of the floating point encoding of continuous optimization functions and the preferred type of solver. Specifically, we discretize several well-known continuous benchmark functions and perform a case study with three classical evolutionary algorithms, all designed to work on different representations. This includes: 
\begin{enumerate}
    \item[i)] a classical $(\mu, \lambda)$-ES~\citep{Beyer.2002}, designed specifically for continuous optimization problems,
    \item[ii)] an Evolutionary Algorithm for integer programming (int-EA)~\citep{Rudolph.1994},
    \item[iii)] a $(\mu + \lambda)$ Genetic Algorithm (GA)~\citep{Hollland.1975} with random integer sampling as mutation operator for discrete optimization problems. 
\end{enumerate} 

Since the latter two algorithms are designed specifically with discrete parameter spaces in mind, we are interested in how $\rho$ impacts the performance of these algorithms compared to the continuous ES.  \\

Additionally, we explore more state-of-the-art algorithms in continuous parameter optimization and focus specifically on the CMA-ES~\cite{Hansen.1996}. We analyze the effect of plateau size on its performance and compare the canonical CMA-ES to a version designed specifically for handling integer variables, the CMA-ES with margin (CMA-ESwM)~\cite{Hamano.2022b}. Here again, we focus on the relation between $\rho$ and the choice of the optimizer, analyzing when it becomes more profitable to use a continuous algorithm or one specifically designed for discrete representations. 

\section{Algorithms included in this study}\label{sec:algorithms}
This section summarizes descriptions of classical and state-of-the-art evolutionary algorithms included in this study. 

\paragraph{Evolution Strategy (ES)} A well-known classic evolutionary algorithm for continuous parameter optimization, first introduced in the 1960s~\cite{rechenberg1965cybernetic, Beyer.2002}. Here, we consider a particular version of ES: a $(\mu, \lambda)$-ES with discrete uniform recombination between two parents selected u.a.r. and mutative self-adaptation of a single endogenous step size parameter $\sigma_i$. The mutation of candidate solutions $\mathbf{x}_{i} \in \mathbb{R}^n$ occurs simultaneously on all object parameters using a multivariate normal distribution: $\mathbf{x}_i' = \mathbf{x}_i + \sigma_i\cdot \mathcal{N}(\mathbf{0}, \mathbf{I})$, and mutation of $\sigma_i$ via scaling with a lognormally distributed random number. We use $\mu=4$ and $\lambda=28$, and initialize for each individual $\sigma_i \sim \mathcal{U}(0, \sqrt{|ub - lb|})$, where $ub$ and $lb$ denote the upper and lower bounds of the domain. 

\paragraph{Evolutionary Algorithm for Integer Programming (int-EA)}  Proposed by Rudolph in 1994~\cite{Rudolph.1994}, this evolutionary algorithm uses a mutation distribution specific for integer search spaces. Following theoretical analysis, it was shown that the maximum entropy mutation distribution for unbounded integer search spaces is $p_k = \frac{p}{2-p}(1 - p)^{|k|}$. A random number with this distribution can be generated by the difference between two independent geometrically distributed random numbers ($G_1$ - $G_2$) parameterized by $p = 1 - m /((1+m)^{1/2} + 1)$. The variance can be controlled via the deviation parameter $m$. Globally, the algorithm is almost completely similar to the $(\mu, \lambda)$-ES, but instead uses mutation with this maximum entropy distribution. The deviation parameter $m$ is encoded identically to $\sigma$  in the ES, as an endogenous strategy parameter and mutated via a lognormal distribution. We again use $\mu=4$ and $\lambda=28$, and initialize $m_i$ equivalent to $\sigma_i$.

\paragraph{Genetic Algorithm (GA)} Originally developed for optimization problems with binary representations, Genetic Algorithms~\cite{Hollland.1975} are yet another branch of evolutionary algorithms that have been widely applied to various optimization problems. GAs are often categorized by the composition of their search operators, \textit{selection}, \textit{mutation}, and \textit{recombination}. Over the years, many sophisticated operators have been developed~\citep{Ye_thesis2022}, but we use a vanilla GA with a simple $(\mu + \lambda)$ deterministic survivor selection mechanism and discrete uniform recombination between two u.a.r. selected parents. Mutation occurs via random uniform resampling with probability $p_m = \frac{1}{n}$ 
for each of the object parameters, which use integer representation, and $\mu=4$ and $\lambda=28$.

\paragraph{Covariance Matrix Adaptation Evolution Strategy (CMA-ES)} A contemporary class of ES, that sets itself apart from traditional ES by its ability to adapt its mutation distribution to an arbitrary multivariate normal distribution $\mathcal{N}(\mathbf{m}, \sigma^2\mathbf{C})$, where $\mathbf{C} \in \mathbb{R}^{n \times n}$ is a covariance matrix and $\mathbf{m}\in \mathbb{R}^{n}$ the center of mass. It consequently can generate correlated mutations and is invariant to (angle-preserving) search space transformations. Many variants have been proposed over the years and it is considered state-of-the-art in real-valued single objective black-box optimization, with many successful applications in real-world optimization. We employ a canonical $(\mu_W, \lambda)$-CMA-ES~\cite{Hansen.1996} with weighted recombination and cumulative step-size adaptation~\cite{Hansen.2016}.  

\paragraph{CMA-ES with margin (CMA-ESwM)} It is known~\cite{Hansen.2011} that the $(\mu_W, \lambda)$-CMA-ES can stagnate when applied to problems with discretized variables, if the variance of the mutation distribution within the CMA-ES becomes, through self-adaptation, smaller than the granularity of the discretization. In other words, when the mutation steps generated by the CMA-ES are smaller than the plateau size~$\rho$, the optimization tends to stick on the plateau. In order to combat this problem, Hamano et.al~\cite{Hamano.2022} proposed a modification to the CMA-ES, the so-called CMA-ES with margin, based on lower-bounding the marginal probabilities of the mutation distribution. This is achieved by introducing a diagonal matrix $\mathbf{A}$ to the mutation distribution $\mathcal{N}(\mathbf{m}, \sigma^2\mathbf{A}\mathbf{C}\mathbf{A}^T)$, and correcting $\mathbf{A}$ and $\mathbf{m}$ such that the probability that the mutation steps are larger than $\rho$ is at least $\alpha$. As a default value for the margin, Hamano et.al~\cite{Hamano.2022} propose $\alpha=\frac{1}{\lambda\:n}$. When $\alpha =0$, no correction is applied, and the CMA-ESwM is equivalent to the original CMA-ES. Results on the bbob-mixint testbed~\cite{Tusar.2019} show that CMA-ESwM outperforms several other methods, particularly at higher dimensions~\cite{Hamano.2022b}. 

\section{Discretization of input variables}
In practice, any real-valued optimization problem can be transformed into a discrete optimization problem by limiting the resolution of the continuous representation. However, this can potentially lead to drastic changes in the search landscape, and the location of the optimum cannot be guaranteed to remain unchanged in the general case. We thus restrict ourselves to problems for which the location of the global optimum $\mathbf{x}^*$ is known, such that we can correct the location of the optimum, for the benefit of this study.  \\

A continuous representation $\mathbf{x} \in [lb,ub]^n \subseteq \mathbb{R}^n$ can be bound to a certain plateau size $\rho$ by applying the transformation $T_{\rho}:\mathbf{x} \to \mathbf{x}_\rho $:
\begin{equation}
    T_{\rho}(\mathbf{x}) = \mathbf{x} - (\mathbf{x}\mod \rho)
\end{equation}
In order to ensure that the global optimum $\mathbf{x}^*$ is included in the discretized space, we apply additional translation: 
\begin{equation}
T_{\mathbf{x}^*}(\mathbf{x}_\rho) = \mathbf{x}_\rho + (\mathbf{x}^* \mod \rho)    
\end{equation}
Since the bounds are not shifted, this translation has the effect that values located on plateaus next to the bounds are shifted outside of the feasible domain, which is corrected as follows:
\begin{equation}
    T_{\mathbf{x}^*}(\mathbf{x}_\rho) = 
    \begin{cases}
        T_{\mathbf{x}^*}(\mathbf{x}_\rho)   &lb < T_{\mathbf{x}^*}(\mathbf{x}_\rho)  < ub \\
        ub  & T_{\mathbf{x}^*}(\mathbf{x}_\rho) > ub \\
        lb  & T_{\mathbf{x}^*}(\mathbf{x}_\rho)  < lb \\
    \end{cases}
\end{equation}
This can cause the plateaus located next to the bounds to span a smaller portion of the underlying domain. 

Considering the original real-valued objective function $f(\mathbf{x})$, for algorithms that use a continuous representation, we can define a discretized objective function with a given plateau size:
\begin{equation}
    f_{\rho}(\mathbf{x}) = f(T_{\mathbf{x}^*}(T_{\rho}(\mathbf{x})))
    \label{problem:con}
\end{equation}

\begin{figure}[!tb]
  \centering
  \includegraphics[width=\columnwidth,trim=1mm 2mm 1mm 1mm,clip]{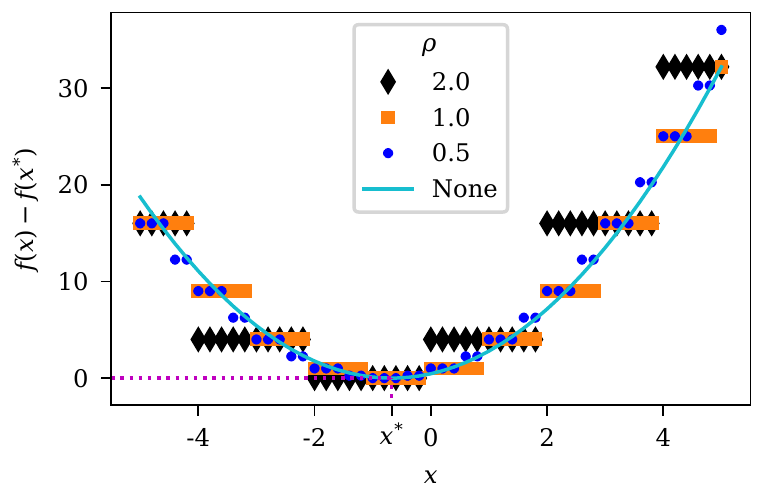}
  \caption{1d plot for the discretized F1 (Sphere) function~$f_{\rho}(\mathbf{x})$ computed according to equation~\ref{problem:con}, within the input-bounds $\mathbf{x}\in [-5, 5]$ for the different plateau sizes~$\rho\in\{\text{None}, 0.5, 1.0, 2.0\}$ with the the global optimum at $\mathbf{x}^*$.}
  \label{fig:discrBBOB}
\end{figure}

Figure~\ref{fig:discrBBOB} illustrates an example of the discretization of a Sphere function in 1D for different plateau sizes~$\rho\in\{0.5, 1, 2\}$. From this figure, we can see that with smaller values of $\rho$ the problem approaches the original continuous version, while for larger values of $\rho$, the number of discrete options decreases. \\

Alternatively, the real-valued domain spanned by the transformation of $T_{\rho}(\mathbf{x})$ can be encoded by an integer space $\mathbf{z} \in [\frac{lb}{\rho},\frac{ub}{\rho}]^n \subseteq \mathbb{Z}^n$, where $\mathbf{x}_\rho = \mathbf{z} \rho$. Algorithms that use integer encoding can then evaluate a real-valued objective function through:
\begin{equation}
    f_{\rho}(\mathbf{z}) = f(T_{\mathbf{x}^*}(\mathbf{z} \rho))
    \label{problem:int}
\end{equation} 

In practice, this allows both integer and continuous solvers to be evaluated on the same discretized version of the original objective function $f(\mathbf{x})$, by parameterizing both $f_{\rho}(\mathbf{z})$ and $f_{\rho}(\mathbf{x})$ with the same $\rho$. This, in turn, allows for a fair comparison, such that we can study the relation between $\rho$ and solver type.

\section{Experimental Setup}\label{sect:exp}

Experiments reported in this study are carried out in the IOHexperimenter environment~\cite{de2021iohexperimenter} which implements the BBOB functions~\cite{Hansen.2009}. We perform experiments on a total of 5 unimodal BBOB problems: F1, F2, F5, F8, and F9, where for each of these problems, instances $\{0, 1, \ldots, 4\}$ are used, with 20 independent runs (for a total of 100 runs per problem). Experiments are performed with plateau sizes~$\rho\in\{\text{None}, 0.001, 0.01, 0.1, 0.5, 1.0, 2.0\}$ and in dimensions $n\in\{2,5,10,20,50\}$. For computational reasons, the budget for each run is $\min(10\,000\cdot n$, $100\,000)$. Additionally, we make use of \textit{hard box-constraints}: points evaluated outside of the domain are considered infeasible and evaluated to $\infty$.  \\

Performance data is collected for each of the algorithms described in Section~\ref{sec:algorithms}, using the aforementioned parameter settings. For the CMA-ESwM, we consider two margin values ~$\alpha\in\left\{\frac{1}{\lambda\:n}, \frac{2}{\lambda\:n}\right\}$, denoted by CMA-ESwM1 and CMA-ESwM2 resp. in the sequel. To measure algorithm performance, we consider the distance in objective space, defined as $\delta_{f^*}=f(\mathbf{x}) - f(\mathbf{x}^*)$, where $f(\mathbf{x})$ is the value of a solution candidate $\mathbf{x}$ provided by the optimization algorithm evaluated on the objective function $f$ and $f^*$ the function value of the optimal solution $\mathbf{x}^*$. We use three derived measures: 
\begin{enumerate}
    \item \emph{Success Rate}, which, for a given budget and targets, returns the proportion of runs which are able to solve the problem within a given budget. BBOB considers a problem "solved" if $\delta_{f^*}$ is less than $10^{-8}$.   
    \item \emph{Expected Running Time} (ERT), which, for a given target, returns the average number of function evaluations, or called average hitting time (AHT), needed to reach that target when the success rate equals $1$. Otherwise, a penalty is assigned based on the number of unsuccessful runs.
    Based on the definition in~\cite{IOHanalyzer}, a simple version of ERT can be represented as $\text{ERT}=\frac{\sum_{i=1}^{r}\min\{t_i(\phi),B\}}{\sum_{i=1}^{r}\mathds{1}\{t_i(\phi)< \infty\}}$, where  $t_i(\phi)$ denotes the function evaluations that one run of the algorithm uses to hit the target value $\phi$, $r$ is the number of independent runs, $B$ is the given budget, and $\mathds{1}(\xi)$ is the indicator function that returns $1$ when the event $\xi$ happens. $t_i(\phi) = \infty$ when the algorithm can not hit the target $\phi$ within the given budget $B$.

    \item \emph{Empirical Cumulative Distribution Function} (ECDF), which records the average success probability among a set of targets $T$. In practice, for each used budget $b$, the ECDF value at $b$ is the fractions of the targets in $T$ that are not better than the best-so-far target the algorithm has obtained. For BBOB functions, $T$ is typically set to 51 logarithmically spaced targets from $10^2$ to $10^{-8}$~\cite{hansen2022anytime}, which we will also use here.
\end{enumerate}

The full experimental setup, including the implementation of each of the described algorithms, has been made available on our Zenodo repository~\cite{reproducibiliyt}. This repository also contains additional reproducibility documents which can be used to re-create the presented figures, as well as additional figures. 

\section{Results}

\subsection{Classical EAs on Discretized Sphere}   
\begin{figure}[!tb]
    \centering
    \includegraphics[width=0.48\textwidth]{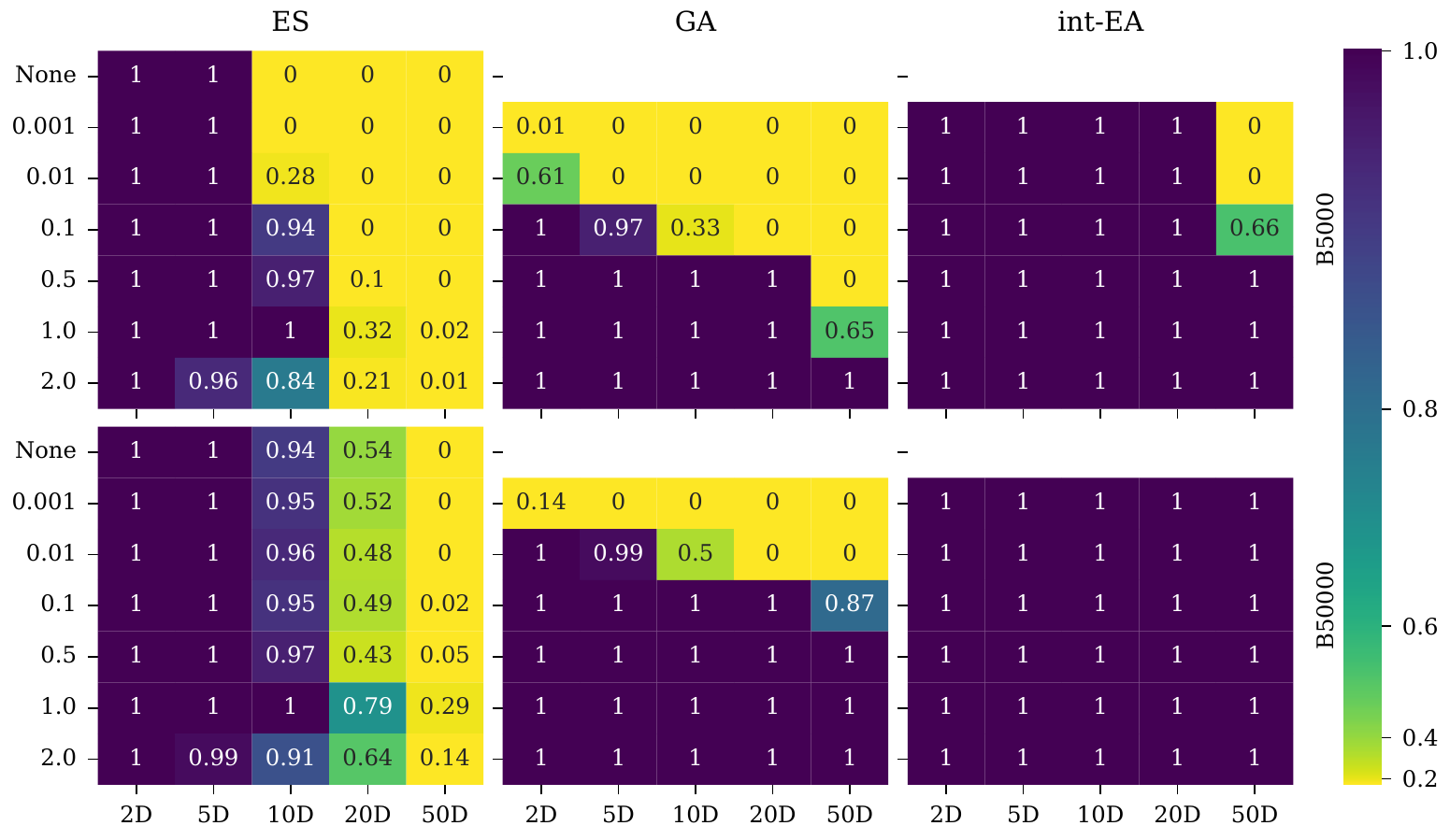}
    \caption{Success rate of runs where the function value reached a target value of $10^{-8}$ by EA, GA, and int-EA on F1 (Sphere) for different dimensions (x-axis) and plateau sizes~$\rho$ (y-axis). Function values are measured after $5\,000$ evaluations (first row) and $50\,000$ evaluations (second row), for a total of $100$ runs per setting.}
    \label{fig:ea_discr_heatmap_f1_suc_2}
\end{figure}

\begin{figure}[!tb]
      \centering
      \includegraphics[width=0.48\textwidth,trim=9mm 13mm 9mm 9mm,clip]{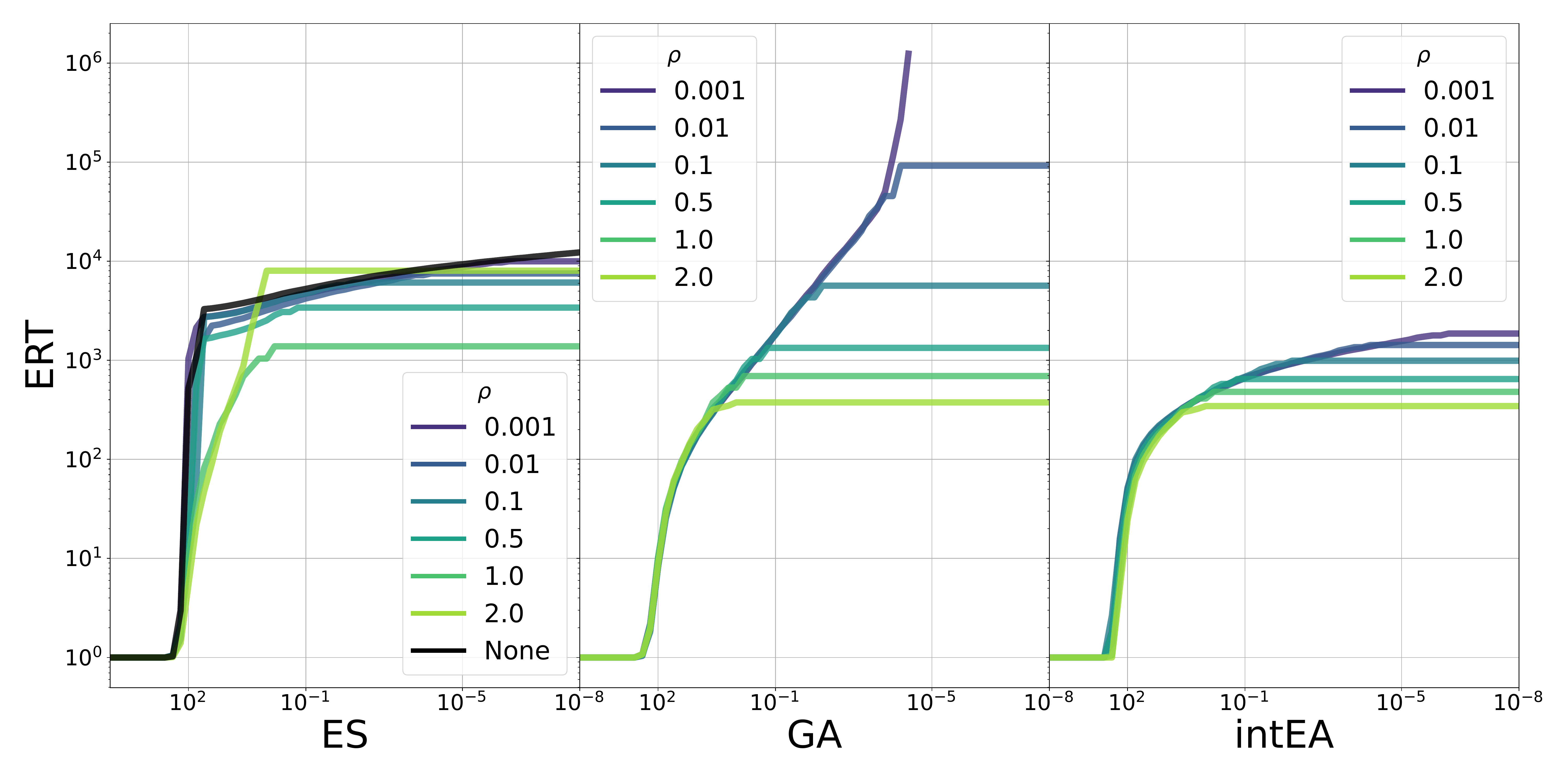}
      \caption{Expected runtime (ERT) of ES, GA, and int-EA on F1 (Sphere) in 10 dimensions for different plateau sizes~$\rho$, calculated from $100$ runs per setting.}\label{fig:ert_grid_f1_10d}
\end{figure}

In this section, we study the impact of the discretization of the Sphere function on the performance of the ($\mu, \lambda$)-ES, GA, and the int-EA. GA and int-EA are specifically designed to handle discrete optimization problems, while ES uses continuous parameter representation. First, we examine the success rate of the algorithms at two cross-cuts of the optimization process from a fixed-budget perspective. In general, since the evaluation budget is the same for all dimensions, higher dimensions have a relatively lower evaluation budget and consequently a lower success rate. Looking at the influence of dimension and plateau size $\rho$, there are some observations to be made from  Figure~\ref{fig:ea_discr_heatmap_f1_suc_2}. For the ES, the success rate decreases with increasing dimensionality and resolution, but the success rate varies only slightly for varying values of $\rho$ for dimensions $<20$. The results obtained within the lower budget of $5\,000$ show that ES struggles with smaller values of $\rho$ in the $10D$ case, where we can clearly observe a lower success rate caused by plateau size. Comparing the results with a budget of $50\,000$ shows that given more budget the ES is able to solve also the more complex problems. Nonetheless, a lower resolution \textit{simplifies} the problem for ES in terms of convergence speed. For the plateau size $0.001$, almost no run of GA is able to solve the problem independently of the dimension. Thus, a lower plateau size \textit{worsens} the performance of GA more than high dimensional problems with a high plateau size. At the same time, the int-EA can solve all tested settings of $\rho$ for the Sphere function, most even with the smaller evaluation budget $5\,000$. Only for $50$ dimensions and the lower plateau sizes, the int-EA needs a higher evaluation budget.  \\

Additionally, we examine the corresponding Expected-Runtime (ERT) of ES, GA, and int-EA on F1 in 10 dimensions for different plateau sizes~$\rho$ (Figure~\ref{fig:ert_grid_f1_10d}), which shows a similar pattern as can be seen in Figure ~\ref{fig:ea_discr_heatmap_f1_suc_2}, but with finer granularity. 
\begin{figure}[!tb]
    \centering
    \includegraphics[width=0.4\textwidth,trim=0mm 1mm 0mm 4mm,clip]{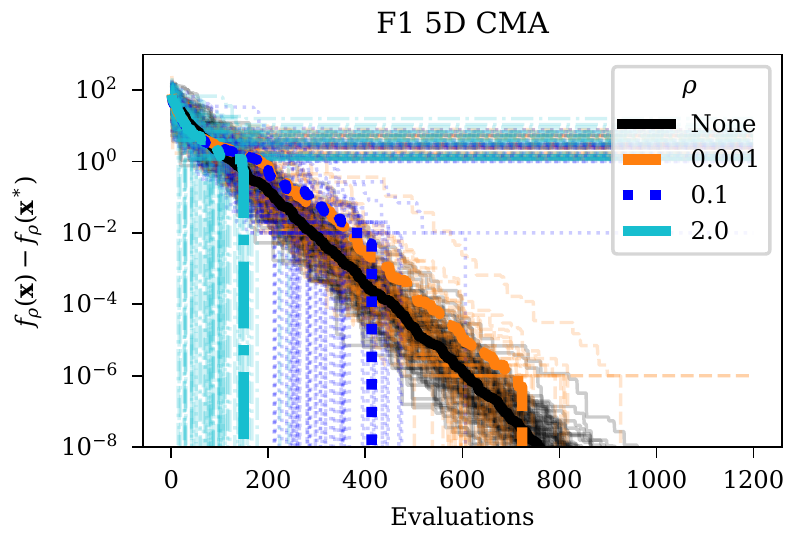}
    \caption{Distance to optimal function value ($\delta_{f^*}$) over evaluations of single CMA-ES runs on F1 (Sphere) in 5 dimensions and different plateau sizes $\rho\in\{\text{None}, 0.001, 0.1, 2.0\}$ for a total of $100$ runs per plateau size and the median (bold line) for each plateau size.}
    \label{fig:cma_runs_f1_5d}
\end{figure}
For ES, the ERT to solve the Sphere function is slightly smaller compared to the continuous case ($\rho = \text{None}$) for all plateau sizes~$\rho$. Therefore, the discretization \textit{simplifies} the problem somewhat for the ES as a solver. For GA and int-EA, a larger plateau size results in a smaller ERT. But unlike the int-EA, for the GA, smaller plateau sizes lead to a much higher ERT. For $\rho = 0.001$ the GA is not expected to solve the problem within a budget of $10^6$ evaluations. Therefore, as the problem becomes more continuous and the number of possible values increases, the performance of GA becomes \textit{excessively degraded} compared to the int-EA. Overall, the ERT of int-EA is the smallest for each plateau size, except for the continuous case where int-EA is not applicable. Thus, to be able to solve continuous optimization problems we deploy CMA-ES in the following.

\subsection{CMA-ES on Discretized Sphere}
CMA-ES is one of the state-of-the-art algorithms for solving continuous optimization problems. To gauge the impact of discretization on CMA-ES, we first examine the canonical $(\mu_W, \lambda)$-CMA-ES without extensions for integer handling on the Sphere function. Figure~\ref{fig:cma_runs_f1_5d} shows the evolution of $\delta_{f^*}$ for each of the individual 100 optimization runs of CMA-ES on the 5~dimensional Sphere function for different plateau sizes $\rho$ as well as the corresponding median values.
Without discretization ($\rho=\text{None}$), CMA-ES is able to solve the Sphere function within about 1000~function evaluations for all 100~runs.
The distance to the optimal function value decreases continuously over the number of evaluations.
When the Sphere function is discretized according to equation~\ref{problem:con}, two phenomena occur (Figure~\ref{fig:cma_runs_f1_5d}). First, for high values of $\rho$, a clear split occurs between runs. One portion of the runs quickly drops down to a $\delta_{f^*}$ below $10^{-8}$, indicating that these runs solved the problem, while another portion of the runs stagnates. This can be explained by considering the fact that the region around the optimal solution is of size $\rho^n$, and any point on this plateau is considered optimal. As such, when the optimization process reaches this plateau, we see a vertical drop in the figure. However, when the plateau is not reached quickly, the CMA-ES can stick in a nearby plateau, with the risk of continuously decreasing its stepsize, lowering the probability of eventually converging. For lower values of $\rho$, both of these effects are reduced, and as such the CMA-ES shows the same behavior as on the continuous problem for a much longer period.
Second, the discretization causes some optimization runs to stagnate for a long time without further improvement of the function value. Thus, CMA-ES becomes stuck on one of the plateaus induced by discretization. This happens due to the step size decreasing to the point where the probability of moving to a new plateau is too low to make progress. \\

\begin{figure}[!tb]
    \centering
    \includegraphics[width=0.48\textwidth]{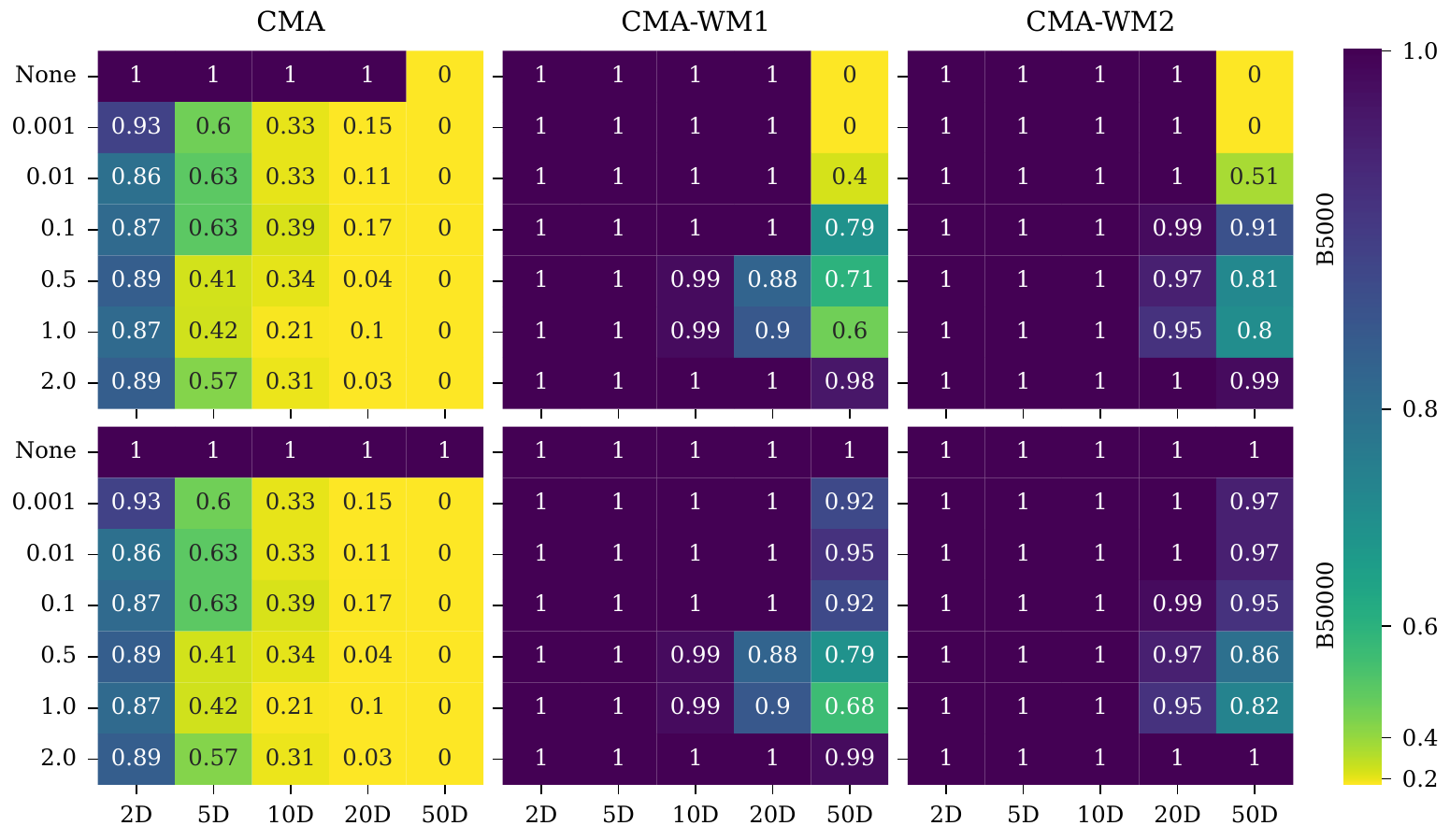}
    \caption{Success rate of runs where the function value reached a target value of $10^{-8}$ by CMA-ES and CMA-ESwM ($\alpha \in \left\{\frac{1}{\lambda\:n}, \frac{2}{\lambda\:n}\right\}$) on F1 (Sphere) for different dimensions (x-axis) and plateau sizes~$\rho$ (y-axis). Function values are measured after $5\,000$ evaluations (first row) and $50\,000$ evaluations (second row), for a total of $100$ runs per setting.}
    \label{fig:cma_discr_heatmap_f1_suc_2}
\end{figure}

\begin{figure}[!tb]
      \centering
      \includegraphics[width=\columnwidth,trim=9mm 14mm 8mm 8mm,clip]{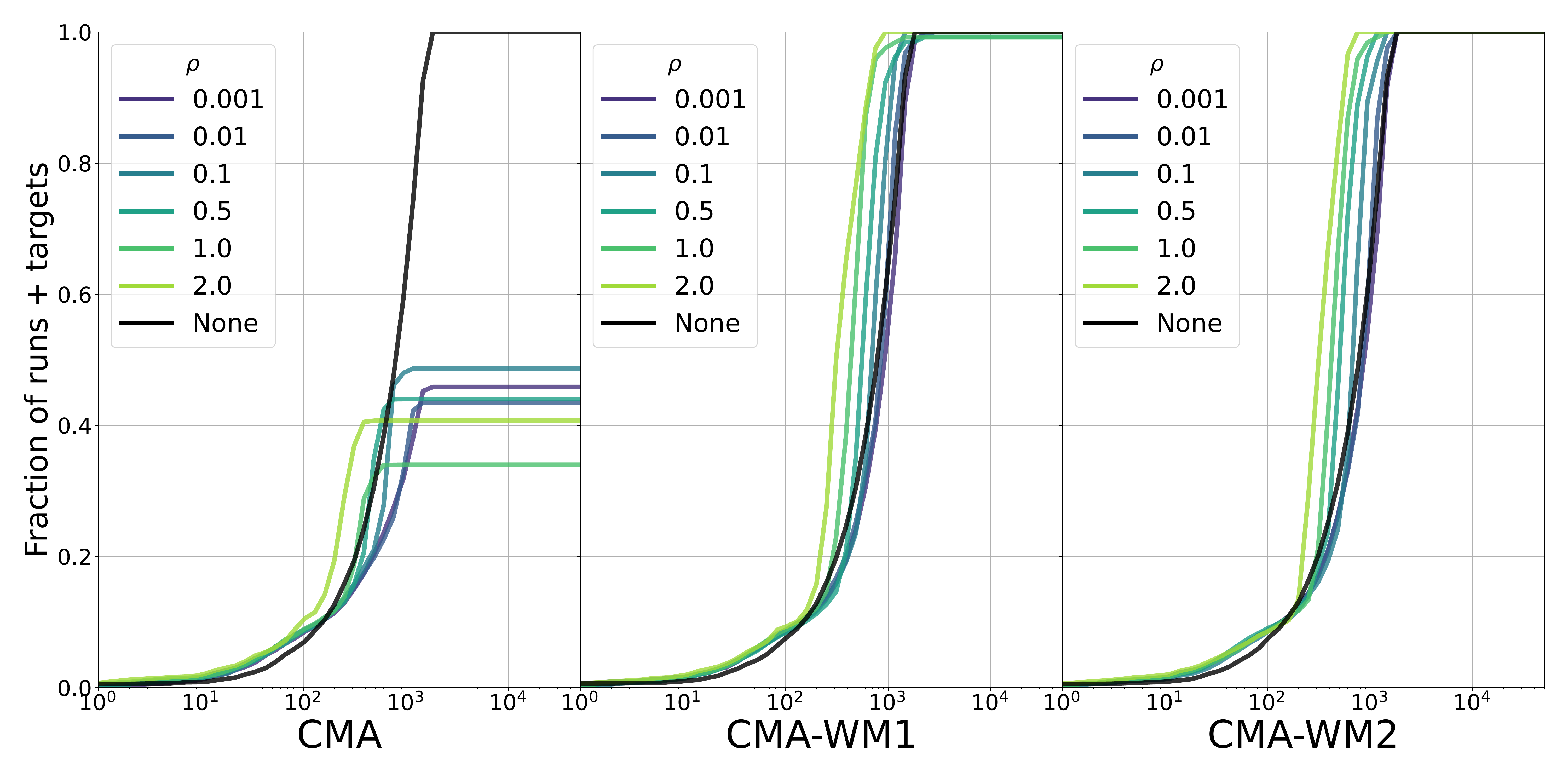}
      \caption{Empirical Cumulative Distribution Function for CMA-ES and CMA-ESwM ($\alpha \in \left\{\frac{1}{\lambda\:n}, \frac{2}{\lambda\:n}\right\}$) on the 10~dimensional F1 (Sphere) for different plateau sizes~$\rho$ that solved the problem within the budget given by the x-axis.}\label{fig:cma_ecdf_grid_f1_10d}
\end{figure}



Subsequently, we investigate the number of runs that solved the problem within a given budget of evaluations. First, we consider the success rate (Section~\ref{sect:exp} for the definition) of the canonical CMA-ES for the two evaluation budgets $5\,000$ and $50\,000$ for different plateau sizes~$\rho$ (Figure~\ref{fig:cma_discr_heatmap_f1_suc_2}). We see that even though CMA-ES manages to solve the continuous Sphere function in $50$ dimensions at all times within $50\,000$ evaluations, adding a certain amount of discretization considerably increases the difficulties faced by the optimizer. The higher the dimension, the more the discretization has a negative impact on the performance of CMA-ES compared to the performance of CMA-ES on the continuous Sphere function ($\rho=\text{None}$). Already in 2D and even for $50\,000$ evaluations, the discretization leads to about 10~\% of optimization runs where CMA-ES cannot solve the problem. The proportion of failed runs increases with the number of dimensions. In the higher dimensions almost no run on the discretized Sphere function finds the solution.  \\



\begin{figure}[!tb]
    \centering
    \includegraphics[width=\columnwidth]{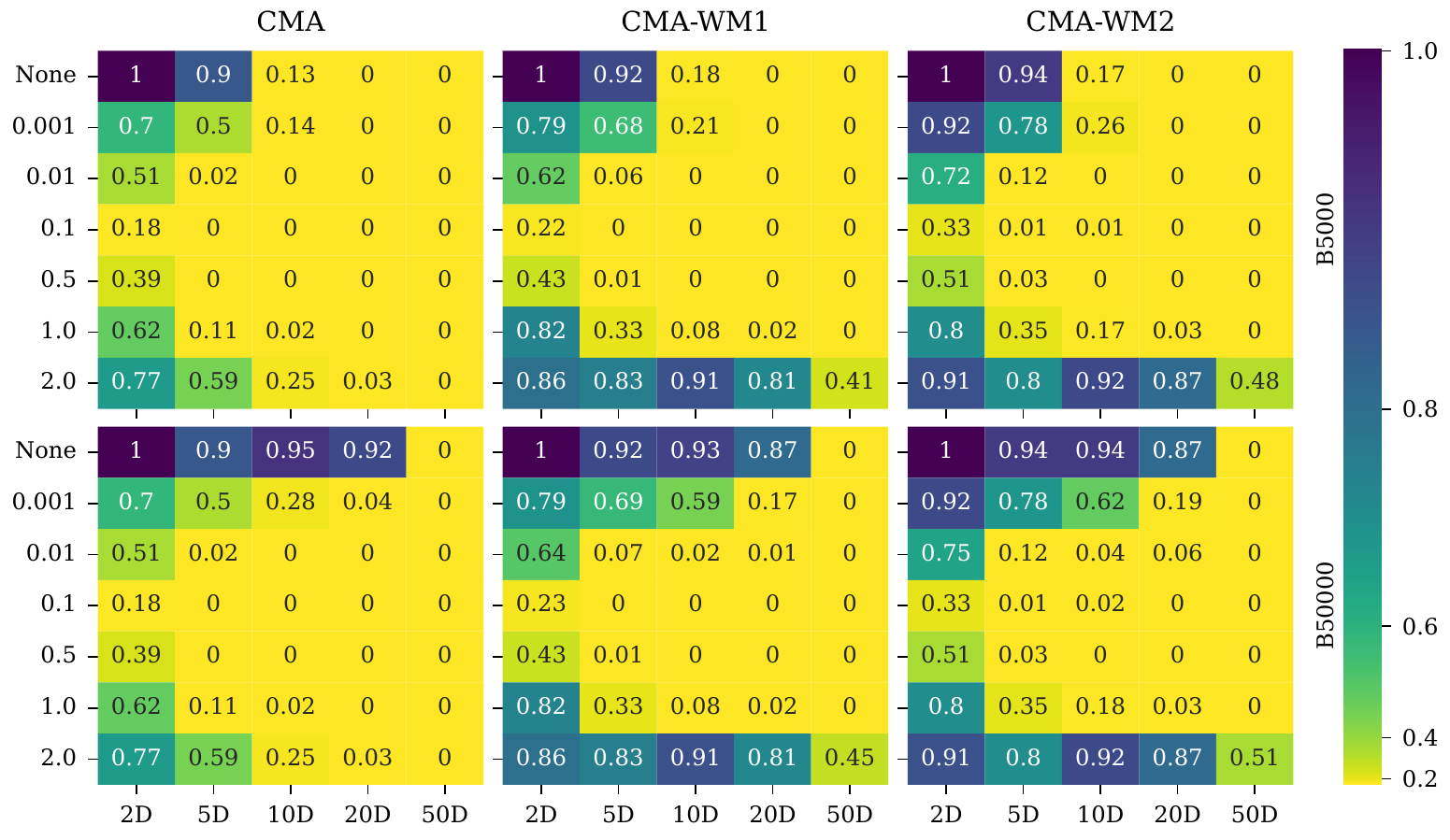}
    \caption{Success rate of runs where the function value reached a target value of $10^{-8}$ by CMA-ES and CMA-ESwM ($\alpha \in \left\{\frac{1}{\lambda\:n}, \frac{2}{\lambda\:n}\right\}$) on F8 (Rosenbrock) for different dimensions (x-axis) and plateau sizes~$\rho$ (y-axis). Function values are measured after $5\,000$ evaluations (first row) and $50\,000$ evaluations (second row), for a total of $100$ runs per setting.}
    \label{fig:cma_heatmap_F8_sucess}
\end{figure}

In the following, we analyze the results of CMA-ES compared to CMA-ES with Margin on the Sphere function (Figure~\ref{fig:cma_discr_heatmap_f1_suc_2}). While we observe that the canonical CMA-ES struggles to handle even small plateau sizes, the extension with the margin improves the performance and outperforms the canonical CMA-ES on all discretized versions of the problems. For a budget of $50\,000$ evaluations almost all runs of CMA-ESwM find the solution regardless of the plateau size. Moreover, when comparing the influence of the margin~$\alpha$ itself on the Sphere function, we observe a minimal improvement by increasing the factor from $\frac{1}{\lambda\:n}$ to $\frac{2}{\lambda\:n}$. 
We also note that, while increasing the evaluation budget has almost no effect on the performance of CMA-ES on the discretized Sphere function, CMA-ESwM improves further. Therefore, in addition to the actual success rate after a certain budget of evaluations, we show in Figure~\ref{fig:cma_ecdf_grid_f1_10d} the success rates over evaluations on the $10$~dimensional Sphere function. Thus, we can get an overview of the way in which CMA-ES and CMA-ESwM converge. Particularly for CMA-ES on the discretized Sphere function ($\rho$ > 0), we see that there is little improvement in the success rate when increasing the budget after $10^3$~evaluations further. Additionally, the evaluations needed to increase the success rate are for all plateau sizes (except of $\rho = 2.0$) higher. \\

With the extension of CMA-ES with the margin, the ECDF on the discretized Sphere function ($\rho$ > 0) is very similar to the continuous case ($\rho$ = None). Therefore, the extension of CMA-ES with the margin does not decrease the convergence speed compared to the runs on the continuous Sphere function. Similar to the int-EA, CMA-ESwM can also benefit from the discretization, the needed evaluation to reach a specific success rate are slightly lower for higher plateau sizes. Therefore, the negative impact of the discretization of the Sphere function on the performance of the canonical CMA-ES turns into a positive impact when the CMA-ES is extended with the margin.

\begin{figure}[!tb]
      \centering
      \includegraphics[width=\columnwidth,trim=9mm 14mm 8mm 8mm,clip]{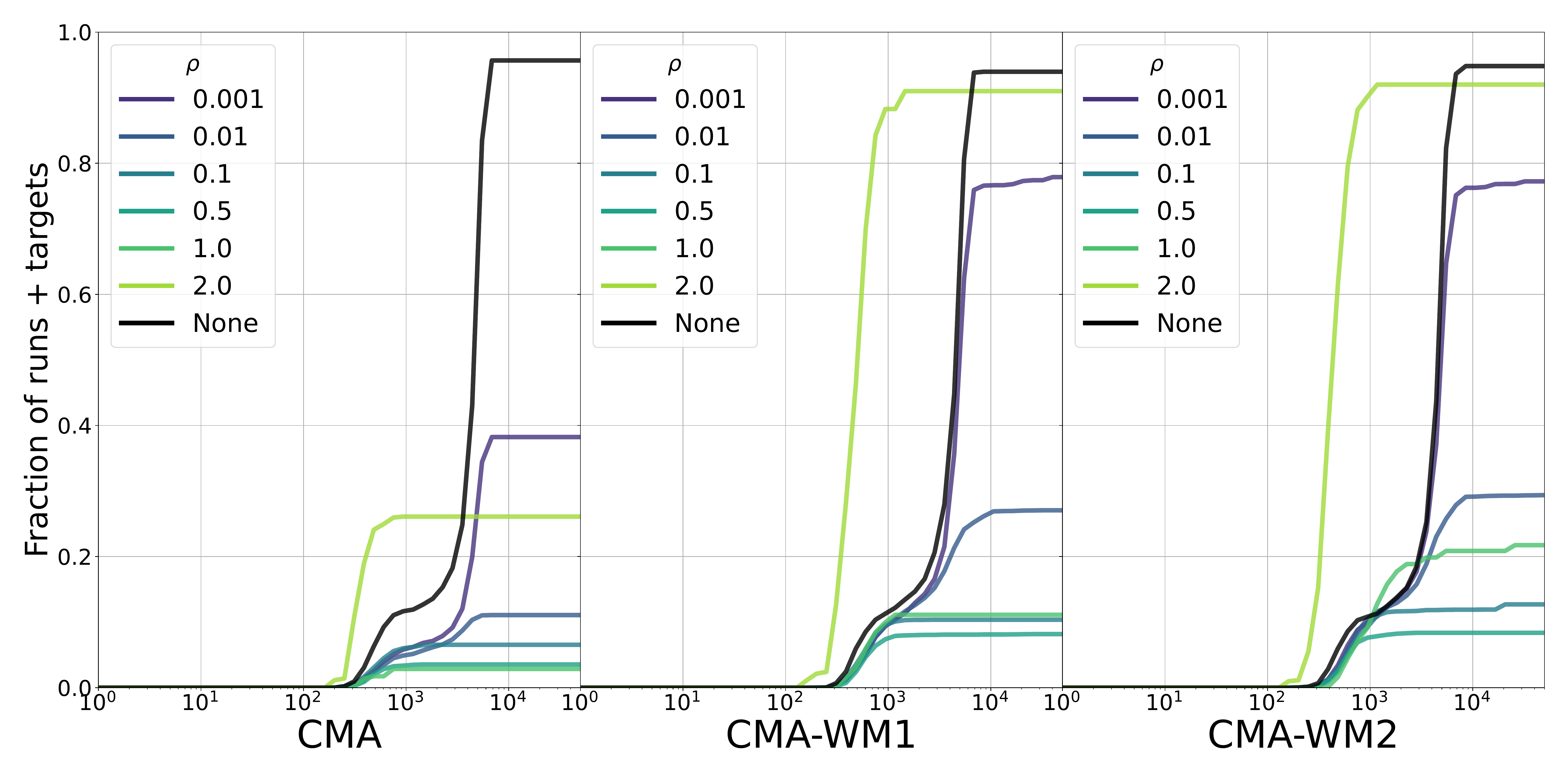}
      \caption{Empirical Cumulative Distribution Function for CMA-ES and CMA-ESwM ($\alpha \in \left\{\frac{1}{\lambda\:n}, \frac{2}{\lambda\:n}\right\}$) on the 10~dimensional F8 (Rosenbrock) for different plateau sizes~$\rho$ that solved the problem within the budget given by the x-axis.}\label{fig:cma_ecdf_grid_f8_10d}
\end{figure}

\subsection{CMA-ES on Discretized Rosenbrock}
In addition to the Sphere function, we also consider the discretized versions of F8, the Rosenbrock function, see  values of the success rate for each of the algorithms across the problem settings in Figure~\ref{fig:cma_heatmap_F8_sucess}. The observed trend differs from the one seen in Figure~\ref{fig:cma_discr_heatmap_f1_suc_2}. While the problem difficulty still obviously increases with dimensionality, the impact of discretization is much less straightforward. Surprisingly, for all versions of CMA-ES, the extreme values for the plateau sizes~$\rho \in \{None, 2\}$ are much more successful than plateau sizes close to $0.1$. Besides the success rate after a certain budget of evaluations, Figure~\ref{fig:cma_ecdf_grid_f8_10d} shows the success rates over evaluations on the $10$~dimensional Rosenbrock function. Particularly for $\rho = 2.0$, we see that in comparision to the other plateau sizes higher success rates are reached earlier. \\

\begin{figure*}[!bt]
      \centering
      \includegraphics[width=\textwidth,trim=12mm 10mm 8mm 8mm,clip]{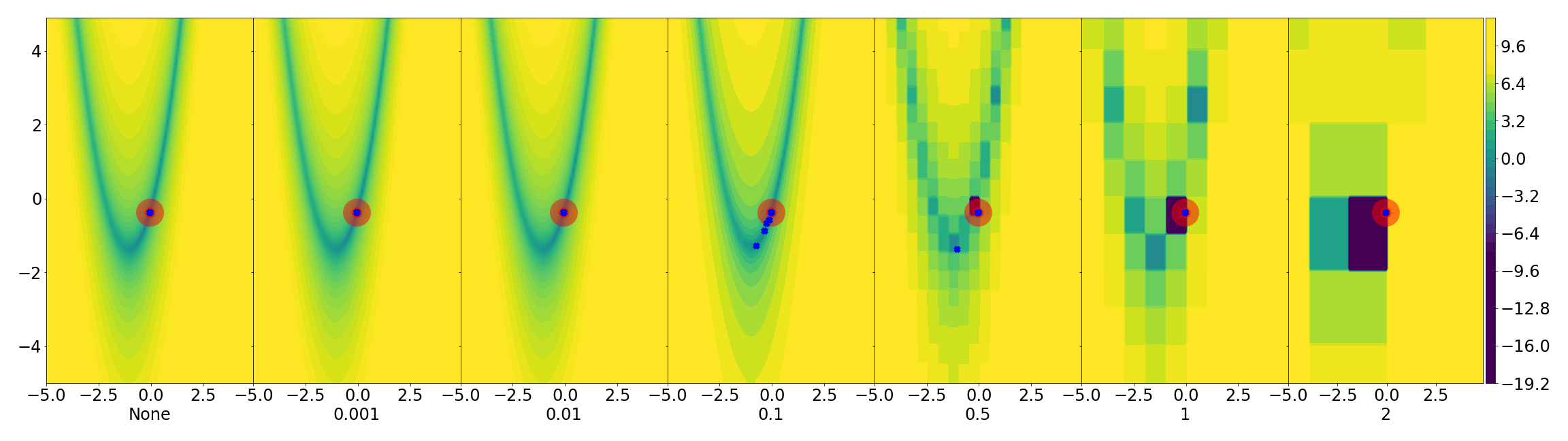}
      \caption{2D plot of the landscape for F8 (Rosenbrock) for different plateau sizes~$\rho\in\{\text{None}, 0.001, 0.01, 0.1, 0.5, 1.0, 2.0\}$. The red circle shows the location of the solution and the blue crosses are the best points found by CMA-ES (indicating the plateau, not the exact location seen by the algorithm itself).
      }\label{fig:F8_landscape}
\end{figure*}

To gain some insight into why the performance of CMA-ES differs so greatly depending on the plateau size, we show the function landscape of the 2-dimensional Rosenbrock problem, for each of the used plateau sizes, in Figure~\ref{fig:F8_landscape}. We observe that the higher plateau sizes seem to remove the main difficulty of constantly having to adapt search directions to follow the ridge. From plateau size~$\rho = 0.5$ upwards, we see why the difficulty is increasing: multiple local optima are being created. This turns the original unimodal problem into a multimodal one, which is more difficult for the CMA-ES to solve. In fact, we can see that several runs of CMA-ES end up in some of these new local optima rather than the global one.

\subsection{All Algorithms on Discretized Functions}
\begin{figure*}
    \centering
    \includegraphics[width=\textwidth]{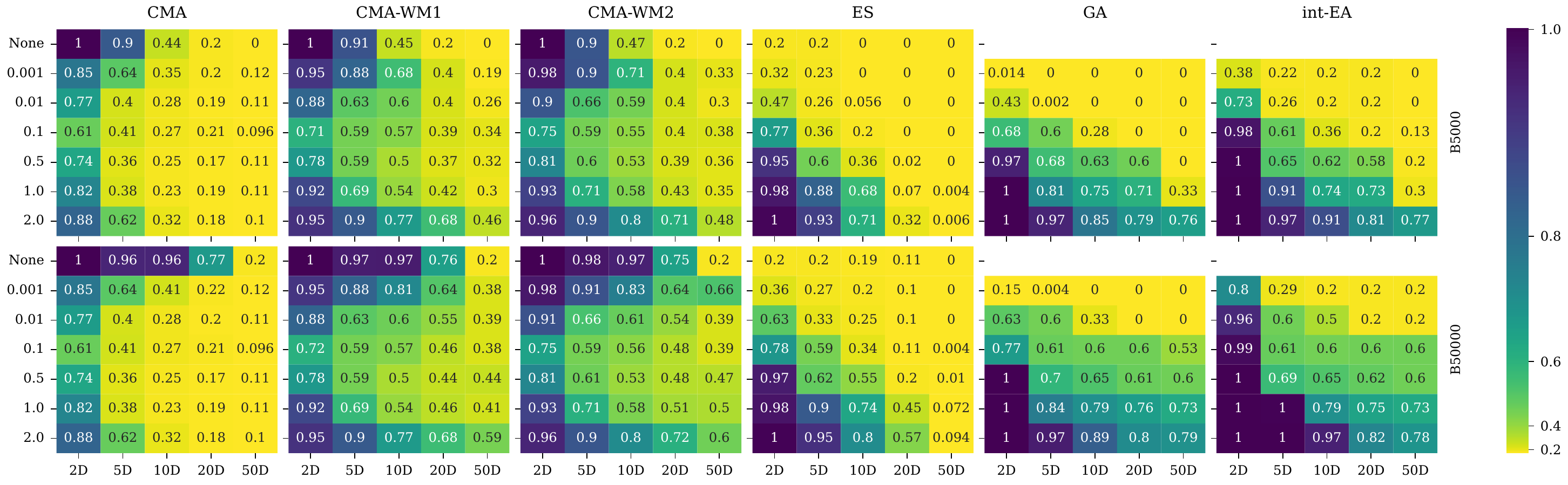}
    \caption{Mean success rate of runs across F1, F2, F5, F8, F9 where the function value reached a target value of $10^{-8}$ by  CMA-ES and CMA-ESwM ($\alpha \in \left\{\frac{1}{\lambda\:n}, \frac{2}{\lambda\:n}\right\}$), EA, GA and int-EA for different dimensions (x-axis) and plateau sizes~$\rho$ (y-axis). Function values are measured after $5\,000$ evaluations (first row) and $50\,000$ evaluations (second row), for a total of $500$ runs per setting.}
    \label{fig:heatmap_Fi_sucess}
\end{figure*}
Next, we compare the performance of the algorithms across all 5~BBOB functions considered. For this, we consider the success rate in a series of $500$ runs ($100$ per function) as a measure of algorithm performance. Figure~\ref{fig:heatmap_Fi_sucess} demonstrates that the CMA-ES with margin~$\alpha=\frac{2}{\lambda\:n}$ outperforms the canonical CMA-ES and the CMA-ES with a smaller margin, while among considered classical algorithms, the int-EA leads to higher success rates compared to ES and the GA. Thus, in the following, we will focus on CMA-ESwM and the int-EA. \\
 
The int-EA performance is decreasing with smaller plateau sizes and higher dimensions, since the number of possible solutions increases.
Due to the extension with the margin CMA-ESwM can handle discrete input-values.
Especially on smaller plateau sizes, where the landscape is similar to the original continuous problem, CMA-ESwM performs better than int-EA.
Thus, despite the int-EA, CMA-ES seems to be able with the adaption of the covariance matrix to adjust the search direction according to the objective landscape. So here CMA-ESwM benefits from its original task of solving continuous problems.
Therefore, we conclude that int-EA is better than CMA-ESwM when the plateau size is not too low ($\rho\leq0.01$). It is of interest to note that at higher plateau sizes ($\rho\geq1.0$) CMA-ESwM can keep up with int-EA. In conclusion, we find that there are only a few settings where the int-EA is especially better than the CMA-ESwM. However, such a conclusion depends on the problem and cannot be generalized. We speculate that to further improve CMA-ESwM, the problem of multimodality arising from discretization must be overcome.





\section{conclusion}
In this paper, we investigated a way to discretize continuous problems, that allows for a fair comparison between continuous and integer optimization algorithms. By discretizing several BBOB functions, we show that when the plateau size is large, a GA is able to consistently outperform an equivalent ES. As the plateau size becomes smaller, this trend reverses, although different integer-based optimization algorithms manage to outperform the ES even for the smallest used plateau size on the Sphere function. In addition to the classical algorithms, we also investigate the impact of discretization on the performance of the CMA-ES. We show that a commonly used version of CMA-ES handles the discretized space poorly, resulting in a stagnation in convergence even on a simple sphere model. This problem is solved with the extension of CMA-ES with the Margin for integer handling, without loss in performance on the continuous version of the problem.

It remains an open question how the differences in performance observed in these experiments translate to other problems. In particular, real-world problems where variables can have different resolutions, are non-separable and multi-modal~\cite{Thomaser.2022} are an interesting area for future research. To gain a deeper understanding of the impact of discretization on the resulting function landscapes, Exploratory Landscape Analysis (ELA)~\cite{Mersmann.2011} could be used to characterize high-level properties such as separability or multimodality.
Furthermore, the resolution of an optimization problem can be seen as a high-level property of an optimization problem. Investigating the relationship between landscape features and resolution could provide insight into whether a clear separation between continuous and discrete optimization is useful for choosing an optimization algorithm, or whether the transition is more fluid.
\begin{acks}
This paper was written as part of the project newAIDE under the consortium leadership of BMW AG with the partners Altair Engineering GmbH, divis intelligent solutions GmbH, MSC Software GmbH, Technical University of Munich, TWT GmbH.
The project is supported by the Federal Ministry for Economic Affairs and Climate Action (BMWK) on the basis of a decision of the German Bundestag.
\end{acks}

\bibliographystyle{ACM-Reference-Format}
\bibliography{references}

\end{document}